\documentclass{article}

\usepackage{spconf,amsmath,graphicx}
\usepackage[T1]{fontenc} 
\usepackage{amsfonts} 
\usepackage{multicol}
\usepackage{multirow}
\usepackage{booktabs}
\usepackage{makecell}
\usepackage{xcolor}
\usepackage{hyperref}
\usepackage{cleveref}
\usepackage{colortbl}
\usepackage{tabulary}
\usepackage{etoolbox}
\usepackage{subfigure}
\usepackage{setspace}
\usepackage{bibspacing}

\definecolor{mygray}{gray}{0.92}
\definecolor{mygray2}{gray}{0.85}

\title{Background clustering  pre-training for Few-shot Segmentation}

\name{Zhimiao Yu, Tiancheng Lin, Yi Xu$^\star$ \thanks{$^\star$ Corresponding author.}}
\address{Shanghai Jiao Tong University, China}
\begin{document}

\maketitle
\begin{abstract}
Recent few-shot segmentation (FSS) methods introduce an extra pre-training stage before meta-training to obtain a stronger backbone, which has become a standard step in few-shot learning. 
Despite the effectiveness, current pre-training scheme suffers from the merged background problem: only base classes are labelled as foregrounds, making it hard to distinguish between novel classes and actual background.
In this paper, we propose a new pre-training scheme for FSS via decoupling the novel classes from background, called Background Clustering Pre-Training (BCPT).
Specifically, we adopt online clustering to the pixel embeddings of merged background to explore the underlying semantic structures, bridging the gap between pre-training and adaptation to novel classes. 
Given the clustering results, we further propose the background mining loss and leverage 
{base classes to guide the clustering process, }
improving the quality and stability of clustering results.
Experiments on PASCAL-5$^i$ and COCO-20$^i$ show that BCPT yields advanced performance. 
Code will be available.


\end{abstract}
\section{Introduction}
In semantic segmentation, most deep leaning methods adopt the fully-supervised learning paradigm and rely heavily on massive pixel-level annotations. However, annotating object masks for large-scale datasets is laborious and expensive. To alleviate this data-hungry nature, few-shot segmentation (FSS) is proposed to segment novel class objects in query images given only a few annotated support images. 

The training process of several recent FSS methods \cite{lu2021simpler, lang2022learning, iqbal2022msanet} includes the pre-training and the meta-training stage. For the first stage, the backbone is pre-trained using the standard supervised learning paradigm on the training dataset of each fold. This pre-training paradigm can obtain a stronger backbone,
and has become a standard step before episodic training in the related field of few-shot image classification~\cite{ye2020few, zhang2020deepemd}.
For the second stage, the episode-based meta-learning strategy is adopted to achieve generalization. Each episode consists of the support set and query set sampled from the training dataset to mimic the few-shot scenarios of novel classes. Existing methods mainly focus on the this stage and try to enhance the generalization ability by fully exploiting the representative information in the limited supports, such as generating more representative prototypes \cite{siam2019adaptive, yang2020prototype, liu2020part, nguyen2019feature, yang2021mining, liu2022dynamic} or proposing better matching mechanism \cite{liu2020dynamic, zhang2021few, hong2021cost, min2021hypercorrelation, shi2022dense}.

\begin{figure}[t]
	\centering

	\begin{minipage}[b]{0.35\textheight}
	\centering
		\includegraphics[width=1\textwidth]{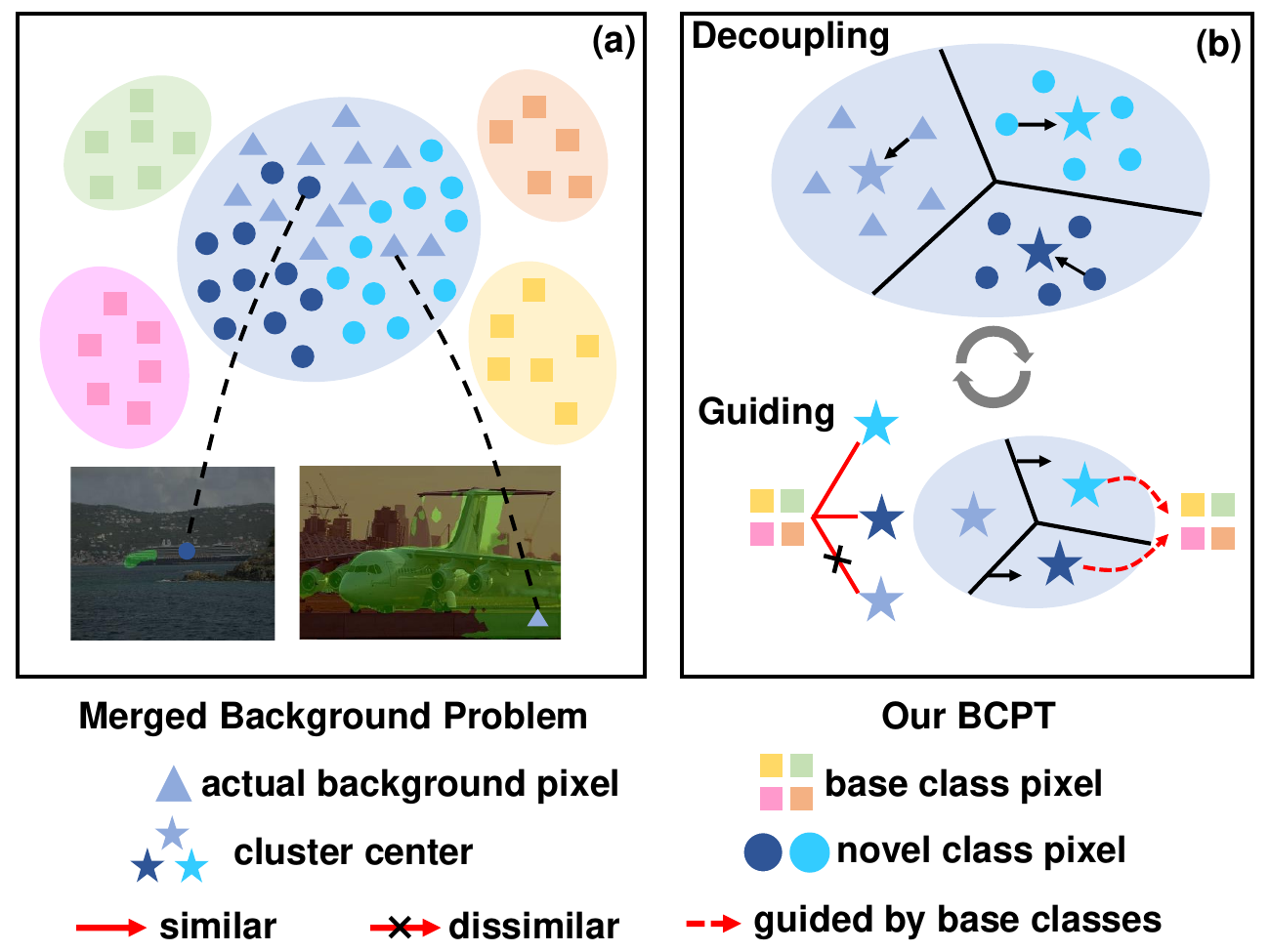}
	\end{minipage}
	\caption{{(a) Current pre-training scheme for FSS suffers from the merged background problem, which yields {inaccurate} FSS predictions: incomplete foreground and incomplete background. (b) Our BCPT explicitly explores the semantic structure of the merged background during pre-training.}}

	\label{fig:fig1}
	\vspace{-0.4cm}
\end{figure}



{Despite the effectiveness of current pre-training scheme, a potential problem when applied to FSS scenario is overlooked by the community.
As shown in Fig.~\ref{fig:fig1}(a), current pre-training suffers from the merged background problem, where novel classes appeared in the training samples are incorrectly learned as background because only annotations of base classes are accessible. As a result, the feature representation of novel classes {presents lowered distinguishability due to undeserved smoothing effects from the background pixels,}
which yields {inaccurate} FSS predictions: 1) The majority of the boat {in the left image} is mistakenly segmented as background. 2) The lower left region of background {in the right image} is mistakenly segmented as aeroplane.}


In our work, we propose background clustering pre-training (BCPT) for few-shot segmentation, which narrows down the gap between pre-training and adaptation to novel classes. As shown in Fig.~\ref{fig:fig1}(b), 
we attempt to decouple {novel classes from background} via exploiting the underlying semantic structure, which is achieved by the unsupervised clustering-based methods~\cite{caron2018deep, zhan2020online}. The multiple cluster centers can capture the diversity of novel classes and actual background.
However, the pseudo-labels from clustering assignments may be unreliable, leading to unstable training process. To alleviate this, we further propose to guide the clustering process with {base classes, considering the the {prior} knowledge that representations of foregrounds ($i.e.$, base and novel classes) share some similarities and differ from that of background~\cite{yang2021mining, xie2022contrastive}.} {Consequently, cluster centers belonging to novel classes could be more separable from those of background via similarity computation.}

The comparative experiments on two prevalent FSS benchmarks show that our BCPT sets the new SOTA on PASCAL-5$^i$~\cite{shaban2017one} and achieves competitive results on COCO-20$^i$~\cite{nguyen2019feature}. In addition, we conduct ablative experiments to validate the design of BCPT.


\label{sec:intro}
\section{Method}

\subsection{Online Clustering for Background}
\label{sec:cluster}
Clustering can assist the learning process to reveal the representative patterns in the data and has achieved success in the area of unsupervised representation learning \cite{caron2018deep, caron2019unsupervised,zhan2020online,  doersch2015unsupervised}. To exploit the semantic structure of the merged background, a straightforward practice is performing the offline clustering algorithm on all background pixel embeddings at the start of each epoch and obtaining the pseudo-labels, $i.e.$, cluster assignments, as the supervision for the next epoch. However, this off-line clustering strategy inevitably permutes the assigned labels in different epochs, making the learning process challenging and unstable. 



To solve this dilemma, we devise an online clustering paradigm to stably assign pseudo-labels for background pixels at each training iteration.  Formally, we define a group of cluster centers, $i.e.$, $ \textbf{P}=\left[ \boldsymbol {p}_1, \cdots,\boldsymbol p_K\right]\in \mathbb{R}^{D\times K}$ 
to estimate the semantic structure of merged background. $D$ and $K$ denote the feature dimension and number of centers, respectively. Given background pixel embeddings $\textbf{I}_{bg}\in \mathbb{R}^{D\times N_{bg}}$ in a training batch, where $N_{bg}$ is the number of background pixels,
we first obtain the similarity matrix $\textbf{S} \in\mathbb{R}^{K\times N_{bg}}$ between pixel embeddings and cluster centers through $\textbf{S}=\textbf{P}^{\text{T}}\cdot \textbf {I}_{bg}$. Then each column of $\textbf{S}$ is transformed to a one-hot vector according to the most similar center of the embedding, which yields the assignment matrix $\textbf{A} \in \mathbb{R}^{K\times N_{bg}}$.
$\textbf{A}$ subsequently aggregates the background pixel embeddings through $\hat{\textbf{P}} =  \textbf {I}_{bg} \cdot \textbf {A}^{\text{T}}$, resulting in $ \hat{\textbf{P}}=\left[ \boldsymbol {\hat{p}}_1, \cdots,\boldsymbol{ \hat{p}}_K\right]\in \mathbb{R}^{D\times K}$. Then, we momentum update each cluster center at each training iteration by:
\begin{equation}
\label{eq:update}
\vspace{-0.05cm}
    \boldsymbol{p} \leftarrow \mu \frac{\boldsymbol{p}}{\Vert \boldsymbol{p} \Vert _2}+ (1-\mu) \frac{\boldsymbol{\hat{p}}}{\Vert \boldsymbol{\hat{p}} \Vert _2},
\end{equation}
where $\mu =0.999$ is a momentum coefficient. With such procedure, the pseudo-labels of background pixels are generated in a stable manner without permutation at different training steps, and the cluster centers evolve continuously by accounting for the cluster assignments. 

\subsection{Network Learning with Cluster Results}
Standard semantic segmentation use a linear layer $\textbf {W}\in \mathbb{R}^{D\times C}$ to project pixel embeddings from high dimensional space to label space, where $C$ is the number of classes. Therefore, one class $c$ is represented by one projection vector $\boldsymbol {w}_c\in \mathbb{R}^D$. For network learning, a pixel embedding is expected to be closer to the projection vector of its corresponding class and further from others of irrelevant classes. We argue this one-to-one learning paradigm, \textit{i.e.}, forcing a pixel embedding of background to a single projection vector, is not suitable in the scenario of FSS pre-training because of the merged background problem.

To alleviate this problem, we extend existing one-to-one learning paradigm to one-to-many, \textit{i.e.}, pulling a pixel embedding closer to its cluster center $\boldsymbol{p}$. Thus we represent background by multiple cluster centers as a group of vectors, and each base class still a single vector. Given a pixel embedding $\boldsymbol{i}$ of background, we define the background mining loss:
\begin{equation}
\vspace{-0.05cm}
    \mathcal{L}_{\text{BM}}^{\boldsymbol{i}} = -\log \frac{\exp(\langle \boldsymbol{i},\boldsymbol{p}_{k} \rangle )}{ \sum _{k'=1}^{K} \exp(\langle \boldsymbol{i},\boldsymbol{p}_{k'} \rangle )},
\end{equation}
where $\langle \;, \rangle$ is the inner product of two vectors, 
and $\boldsymbol{p}_k$ is determined by the assignment matrix $\textbf{A}$. Compared with standard segmentation, the learning objective for different pixel embeddings belonging to background is consistent with the semantic structure explored by online clustering,  which shapes the feature space more accurately.

For a pixel embedding $\boldsymbol{j}$ of base classes, we utilize the standard cross-entropy loss since we are accessible to its ground truth $c$:
\begin{equation}
\vspace{-0.05cm}
    \mathcal{L}_{\text{Base}}^{\boldsymbol{j}} = -\log \frac{\exp (\langle \boldsymbol j\;, \boldsymbol{w}_c \rangle)}{\sum _{c'} \exp (\langle \boldsymbol j\;, \boldsymbol{w}_{c'} \rangle)}. 
\label{eq:dist}
\end{equation}
where $\boldsymbol{w}_c$ is the corresponding projection vector of class $c$. Given pixel embeddings $\textbf{I}^{base}$ and $\textbf{I}^{bg}$ of base classes and background in a batch, we minimize the combinational loss to train the network:
\begin{equation}
\vspace{-0.05cm}
        \mathcal{L}_{\text{total}}=\sum _{\boldsymbol{j}\in \textbf{I}^{base}}\mathcal{L}_{\text{Base}}^{\boldsymbol{j}} + \alpha \sum _{\boldsymbol{i}\in \textbf{I}^{bg}}\mathcal{L}_{\text {BM}}^{\boldsymbol{i}}.
\end{equation}
where pixel embeddings $\textbf{I}^{base}$ and $\textbf{I}^{bg}$ are generated by the last layer of the backbone, and $\alpha =0.1$ is utilized to balance the loss scale.


\subsection{Online Clustering with Guidance}
The online clustering strategy introduced in \S{2.1}  may lead to inferior results, due to two factors: (1) The randomly initialized cluster centers are not able to comprehensively retrieve affiliated pixel embeddings at the beginning of training. (2) The simple clustering and updating design may suffer from trivial solutions. To tackle these challenges, we attempt to connect the learning of novel classes with that of base classes, which receive accurate and sufficient supervisions.
According to~\cite{yang2021mining, xie2022contrastive}, {we have the prior knowledge that representations of foregrounds (i.e., base and novel classes) share some similarities and differ from that of background. From this perspective, we use base classes to guide the clustering procedure and thus achieve more accurate clustering results.}



First, we perform $k$-means algorithm on projection vectors $\{ \boldsymbol{w}_c\}$, {which abstract the representation of base classes,} to generate $K-1$ guidance vectors:
\begin{equation}
\vspace{-0.05cm}
    \textbf{G} \leftarrow k\text{-means}(\{\boldsymbol{w}_c \}),
\end{equation}
where $\textbf{G}=\left[ \boldsymbol {g}_1, \cdots,\boldsymbol{g}_{K-1}\right]\in \mathbb{R}^{D\times (K-1)}$. Second, we map $K-1$ guidance vectors to $K$ cluster centers and denote the mapping matrix as $\textbf{M}\in \mathbb{R}^{(K-1)\times K}$, where each row $\boldsymbol{m}_i\in \mathbb{R}^{K}$ is the one-hot assignment vector of $\boldsymbol{g}_i$ over $K$ cluster centers. The mapping matrix is achieved by maximizing the similarity score between guidance vectors and cluster centers:
\begin{equation}
\vspace{-0.05cm}
\label{eq:M}
\begin{aligned}
    \mathop{\max}_{\textbf{M}} \text{Tr}(\textbf{M}^{\text{T}} \cdot \textbf{G}^{\text{T}} \cdot \textbf{P}), \quad \quad \quad \quad\\
    s.t.\quad \textbf{M}\in \{0,1 \}^{(K-1)\times K}, \boldsymbol{1}^{K} \cdot  \textbf{M}^{\text{T}}=\boldsymbol{1}^{K-1},
\end{aligned}  
\end{equation}
where $\boldsymbol{1}^K$ denotes the vector of all ones of $K$ dimensions. 
{Finally, we momentum update the cluster centers with assigned guidance vectors {in a similar manner to} Eq.~\ref{eq:update}:
\begin{equation}
\vspace{-0.05cm}
\label{eq:g}
    {\boldsymbol{p} \leftarrow \mu \boldsymbol{p}+(1-\mu)\frac{\boldsymbol{\hat{g}}}{\Vert \boldsymbol{\hat{g}} \Vert _2}},
\end{equation}
where $\boldsymbol{\hat{g}}$ indicates the sum of guidance vectors assigned to $\boldsymbol{p}$.
Since $K-1<K$, at least one cluster center will not be assigned with any guidance vector and not be updated. 
Meanwhile, updated cluster centers are inclined to retrieve novel class pixel embeddings from background pixel embeddings, leading to {more distinguishable} clustering results.
The above guidance steps are performed at the beginning of each training iteration.}

\begin{table*}[t]
	\centering
		\caption{Performance on PASCAL-5$^i$ and COCO-20$^i$. Bold and underlined numbers highlight the best and second best performance for each backbone, respectively.}
	\resizebox{0.87\linewidth}{!}{
		\renewcommand{\arraystretch}{1.1}
		\begin{tabular}{p{1.3cm}>{\hfill}p{3.7cm}p{1.7cm}|p{1.0cm}<{\centering}p{1.0cm}<{\centering}p{1.0cm}<{\centering}p{1.0cm}<{\centering}p{1.0cm}<{\centering}|p{1.0cm}<{\centering}p{1.0cm}<{\centering}p{1.0cm}<{\centering}p{1.0cm}<{\centering}p{1.0cm}<{\centering}}
			\hline
            \multicolumn{13}{c}{\cellcolor{mygray2}\textit{PASCAL-5}$^i$} \\
            \hline
			\multirow{2}{*}{Backbone} & \multirow{2}{*}{FSS Method} &\multirow{2}{*}{Pre-train} & \multicolumn{5}{c|}{1-shot mIoU}           & \multicolumn{5}{c}{5-shot mIoU}            \\ \cline{4-13} 
			&   &         & Fold-0 & Fold-1 & Fold-2 & Fold-3 & Mean  & Fold-0 & Fold-1 & Fold-2 & Fold-3 & Mean  \\ \hline
			\multirow{7}{*}{VGG16}    
			& FWB (ICCV'19)\cite{nguyen2019feature} &  \multirow{4}{*}{ImageNet}      & 47.00  & 59.60  & 52.60  & 48.30  & 51.90 & 50.90  & 62.90  & 56.50  & 50.10  & 55.10 \\
			& PFENet (TPAMI'20)\cite{tian2020prior} &   & 56.90  & {68.20}  & 54.40  & 52.40  & 58.00 & 59.00  & 69.10  & 54.80  & 52.90  & 59.00 \\
			& HSNet (ICCV'21)\cite{min2021hypercorrelation} &     & 59.60  & 65.70  & 59.60  & 54.00  & 59.70 & \underline{64.90}  & 69.00  & 64.10  & 58.60  & 64.10 \\
			& DPCN(CVPR'22)~\cite{liu2022dynamic} &    & 58.90 & \underline{69.10}& 63.20& \underline{55.70}& 61.70 & 63.40& 70.70& 68.10& 59.00& 65.30 \\	
			\cline{2-13} 
			& ASPP PFENet & ImageNet  & 59.45 &	66.65&	63.22&	54.77&	61.02 & 63.39&	70.11	&67.66&	62.51&	65.92 \\
			& ASPP PFENet&  Standard & \underline{60.02}&	68.67&	\underline{63.76}&	55.00&	\underline{61.86} & 64.02&	\underline{71.51}&	\underline{69.39}&	\underline{63.55}&	\underline{67.12} \\
			& \cellcolor{mygray}ASPP PFENet & \cellcolor{mygray}{BCPT(ours)} & \cellcolor{mygray}\textbf{60.77}  & \cellcolor{mygray}\textbf{69.96}  & \cellcolor{mygray}\textbf{64.28}  & \cellcolor{mygray}\textbf{56.60}  & \cellcolor{mygray}\textbf{62.90} & \cellcolor{mygray}\textbf{65.21}  & \cellcolor{mygray}\textbf{72.68}  & \cellcolor{mygray}\textbf{70.81}  & \cellcolor{mygray}\textbf{64.81}  & \cellcolor{mygray}\textbf{68.38} \\ \hline
			\multirow{8}{*}{ResNet50} & PFENet(TPAMI'20)~\cite{tian2020prior} &  \multirow{4}{*}{ImageNet}              & 61.70 & 69.50 & 55.40 & 56.30 & 60.80 & 63.10 & 70.70 & 55.80 & 57.90 & 61.90  \\
			& HSNet(ICCV'21)~\cite{min2021hypercorrelation} & &64.30& 70.70& 60.30 &60.50& 64.00 &70.30 &73.20 &67.40 &\underline{67.10}& 69.50 \\
			&DCAMA(ECCV'22)~\cite{fan2022self} & & \underline{67.50}& \textbf{72.30}& 59.60& 59.00& 64.60&  \underline{70.50}& \underline{73.90}& 63.70& 65.80& 68.50 \\
			& DPCN(CVPR'22)~\cite{liu2022dynamic}&  & 65.70 & \underline{71.60} & \textbf{69.10} & \underline{60.60} & \underline{66.70} & 70.00 & 73.20 & \textbf{70.90} & 65.50 & \underline{69.90} \\ \cline{2-13} 
			& ASPP PFENet & ImageNet  &  64.42 & 70.81 & 63.64 & 57.98 & 64.21 & 67.10&	72.31&	63.64&	63.19&	66.56\\
						& ASPP PFENet & Standard  &  64.79 &	71.46&	65.94&	59.51&	65.36 & 67.59&	73.21&	67.63&	65.97&	68.60\\
			& \cellcolor{mygray}ASPP PFENet & \cellcolor{mygray}{BCPT(ours)} & \cellcolor{mygray}\textbf{68.73}  & \cellcolor{mygray}{71.57}  & \cellcolor{mygray}\underline{65.69}  & \cellcolor{mygray}\textbf{61.22}  & \cellcolor{mygray}\textbf{66.87} & \cellcolor{mygray}\textbf{73.20}  & \cellcolor{mygray}\textbf{74.25}  & \cellcolor{mygray}\underline{69.38}  & \cellcolor{mygray}\textbf{67.50}  & \cellcolor{mygray}\textbf{71.21} \\ 
			\hline
            \multicolumn{13}{c}{\cellcolor{mygray2} \textit{COCO-20}$^i$} \\
            \hline
            \multirow{6}{*}{ResNet50} & HSNet (ICCV'21)\cite{min2021hypercorrelation} &  \multirow{3}{*}{ImageNet}              & 36.30& 43.10 &38.70& 38.70& 39.20 &  43.30 & 51.30 &48.20 &45.00& 46.90   \\
            & DCAMA(ECCV'22)~\cite{fan2022self} &         & \underline{41.90}& 45.10 &\textbf{44.40}& 41.70& \underline{43.30} &  \underline{45.90} & 50.50 & 50.70 &46.00& 48.30 \\
            & DPCN(CVPR'22)~\cite{liu2022dynamic} &    & \textbf{42.00} & 47.00 & 43.20 & 39.70 & 43.00 & \textbf{46.00}& \textbf{54.90}& \underline{50.80}& 47.40& \underline{49.80} \\	
            \cline{2-13} 
            & ASPP PFENet &  ImageNet & 37.11&	47.47&	43.20&	41.28&	42.27& 43.25&	52.94&	48.58&	48.01&	48.20      \\
            & ASPP PFENet &  Standard       & 37.57 & \underline{48.21} &44.13& \textbf{42.68}& 43.15 & 44.89&	52.85&	50.30&	\underline{48.44}&	49.12\\
            & \cellcolor{mygray}{ASPP PFENet} &    \cellcolor{mygray}{BCPT(ours)}     & \cellcolor{mygray}{37.50} & \cellcolor{mygray}{\textbf{51.15}} & \cellcolor{mygray}{\underline{44.38}} & \cellcolor{mygray}{\underline{42.33}} & \cellcolor{mygray}{\textbf{43.84}} & \cellcolor{mygray}{45.47}&	\cellcolor{mygray}{\underline{53.98}}&	\cellcolor{mygray}{\textbf{50.94}}&	\cellcolor{mygray}{\textbf{49.01}} &	\cellcolor{mygray}{\textbf{49.85}} \\ 
            
            \hline
	\end{tabular}}
	\vspace{-0.3cm}
	\label{tab:pascal}
\end{table*}



\begin{table}[t]
\caption{Non-parametric performance on PASCAL-5$^i$.}
\label{tab:ab_nopara}
\small
\centering
\resizebox{0.85\linewidth}{!}{
\tabcolsep 0.08in\begin{tabular}{l|ccccc}
\toprule[1pt]
\multirow{2}{*}{Pre-train}  & \multicolumn{5}{c}{1-shot mIoU}   \\
 &   Fold0 & Fold1 & Fold2 & Fold3 & Mean \\

\hline
ImageNet   & 27.04 & 38.29 & 36.40 &33.88 & 33.90 \\

Standard    &28.65	&40.43	&{43.33}&	{36.85}&	37.32 \\

BCPT (ours)   &\textbf{38.52}	&\textbf{47.08}&	\textbf{46.01}&	\textbf{38.65}&	\textbf{42.57}  \\

\bottomrule[1pt]
\end{tabular}}
\vspace{-0.3cm}
\end{table}

\begin{table}[t]
\caption{Ablation studies of each component on PASCAL-5$^i$. `Mean' denotes mean mIoU over 4 folds under 5-shot setting.}
\label{tab:ab_comp}
\small
\centering
\resizebox{0.55\linewidth}{!}{
\setlength{\tabcolsep}{1.5mm}{

\begin{tabular}{cc|cc}
\toprule[1pt]
BMC  & OCG  & Mean & FB-IoU \\
\hline
 &   &  68.60 & 79.55\\
\checkmark  &   &	70.15 & 80.44 \\
\checkmark   & \checkmark  &	\textbf{71.21}& \textbf{81.16} \\

\bottomrule[1pt]
\end{tabular}}}
\vspace{-0.3cm}
\end{table}

\begin{table}[t]
\caption{ Ablation on cluster numbers $K$ on PASCAL-5$^i$.}
\label{tab:ablation2}
\small
\centering
\resizebox{0.85\linewidth}{!}{
\tabcolsep 0.08in\begin{tabular}{c|ccccc|c}
\toprule[1pt]
\multirow{2}{*}{$K$}  & \multicolumn{5}{c|}{5-shot mIoU}  &\multirow{2}{*}{FB-IoU}   \\
& Fold0 & Fold1 & Fold2 & Fold3 & Mean  \\
\hline
2  & 72.89 &	74.95&	67.15&	\textbf{68.87}&	70.97 & 81.09\\
3  & \textbf{74.05}&	\textbf{75.54}&	68.30&	66.41&	71.08 &81.15 \\
6  & 73.20&	74.25&	\textbf{69.38}&	67.50&	\textbf{71.21} &\textbf{81.16} \\

\bottomrule[1pt]
\end{tabular}}
\vspace{-0.3cm}
\end{table}

 \begin{figure}[t]
	\centering

	\begin{minipage}[b]{0.33\textheight}
	\centering
		\includegraphics[width=1\textwidth]{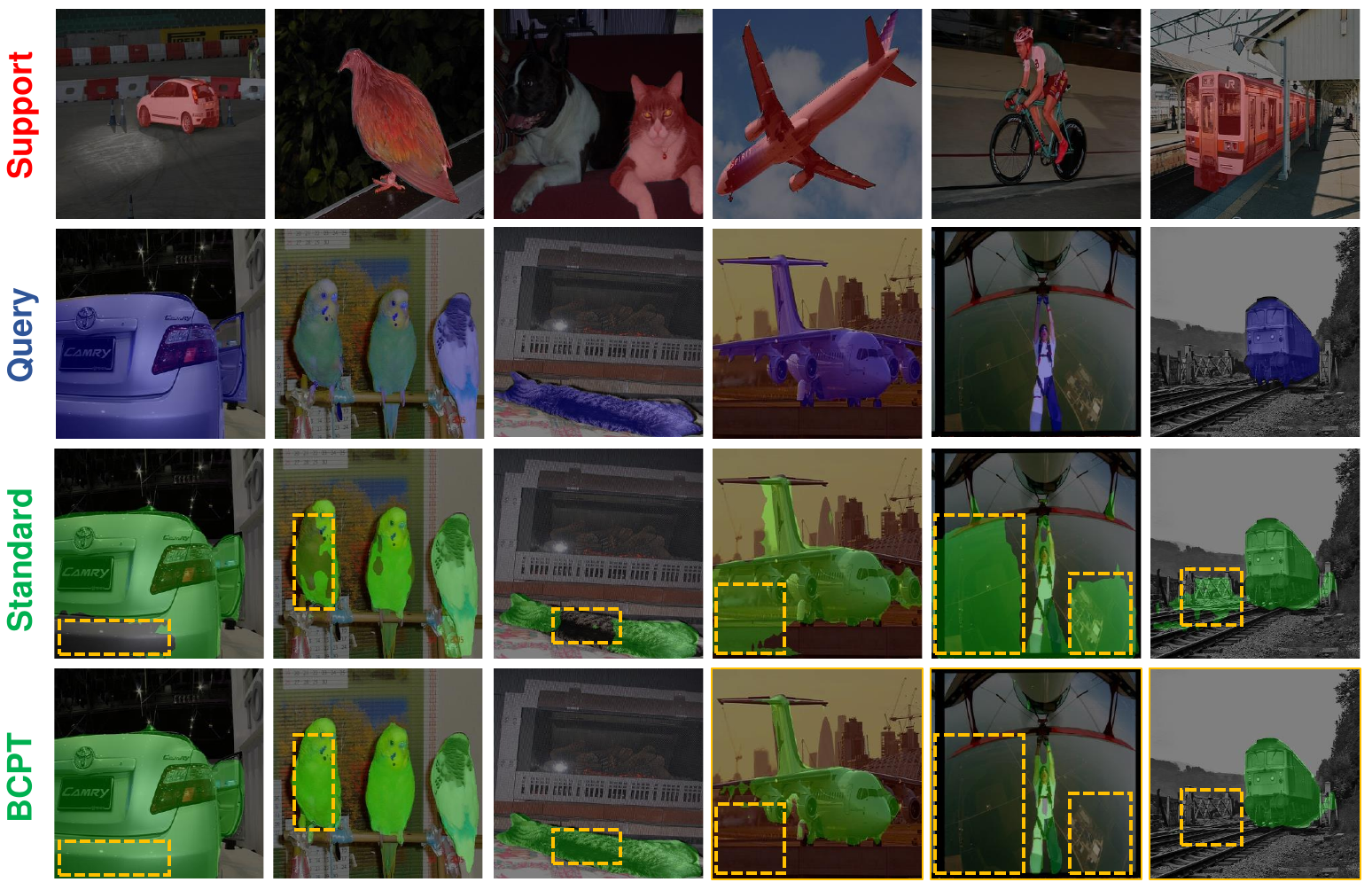}
	\end{minipage}
	\caption{{{Visual comparisons} on PASCAL-5$^i$. The main differences between BCPT and "Standard" are highlighted with yellow boxes.}}

	\label{fig:fig2}
\vspace{-0.4cm}
\end{figure}

\section{Experiments}
\subsection{Setup}
\textbf{Dataset and metrics.}
We evaluate the proposed method on two widely-used FSS benchmarks: PASCAL-5$^i$~\cite{shaban2017one} and COCO-20$^i$~\cite{nguyen2019feature}. We adopt mean intersection-over-union (mIoU) and foreground-background IoU (FB-IoU) as the evaluation metrics.


\noindent \textbf{Pre-training.}
For comparison, we adpot different methods to pre-train the backbone. \textbf{"ImageNet"} means the backbone is initialized with ImageNet~\cite{deng2009imagenet} pre-trained weights.
\textbf{"Standard"} indicates the backbone is pre-trained with the standard supervised segmentation paradigm on each fold, where only annotations of base classes are accessible.
\textbf{"BCTP"} is the proposed pre-training method.
Following~\cite{lang2022learning}, we train the network with an SGD optimizer on PASCAL-5$^i$ for 100 epochs and COCO-20$^i$ for 20 epochs.


\noindent \textbf{Meta-training.}
Meta-training is performed after pre-training. In this stage, the backbone is freezed and extracts the features of input images. 
We meta-train the FSS network in an episodic manner with an SGD optimizer on PASCAL-5$^i$ for 200 epochs and COCO-20$^i$ for 50 epochs.

\noindent\textbf{Network architecture.}
We use VGG-16~\cite{simonyan2014very} and ResNet-50~\cite{he2016deep} as the backbone networks.
Following \cite{lang2022learning, sun2022singular}, we replace FEM module on PFENet~\cite{tian2020prior} with ASPP module to obtain the FSS network, named as ``ASPP PFENet".


\subsection{Main Results}
We compare the performance of the proposed BCPT with SOTA methods on PASCAL-5$^i$ and COCO-20$^i$.

\noindent\textbf{PASCAL-5$^i$.}
In Table~\ref{tab:pascal}, on both VGG-16 and ResNet-50, BCPT outperforms "ImageNet" by a large margin, while "Standard" brings limited improvements. With ResNet-50, BCPT surpasses "Standard" by 1.51\% and 2.61\% mIoU under 1-shot and 5-shot setting respectively. With the help of BCPT, ASPP PFENet achieves 3.08\% and 1.31\% mIoU improvement over previous SOTA DPCN~\cite{liu2022dynamic} under 5-shot setting with VGG-16 and ResNet-50 respectively. To further verify the generalization ability of BCPT, we evaluate the performance 
in a non-parametric manner. Following~\cite{tian2020prior}, we take the output of the last layer of conv5\_x in ResNet-50 to generate the prior mask, and convert it to the segmentation mask with the same threshold. We report the performance in Table~\ref{tab:ab_nopara}. BCPT significantly outperforms "Standard" and "ImageNet" by 5.25\% and 8.67\% mIoU, which proves that BCPT can obtain more discriminative feature representations.

\noindent\textbf{COCO-20$^i$.}
As presented in Table~\ref{tab:pascal}, BCPT achieves 0.69\% and 0.73\% mIoU  improvement over ``Standard" with ResNet-50 backbone under 1-shot and 5-shot setting respectively. Moreover, BCPT obtains comparable or better results compared with previous SOTA methods.

 
\noindent\textbf{Qualitative Results.}
{In Fig.~\ref{fig:fig2}, we report some {visual comparisons} between BCPT and "Standard" under 1-shot setting on PASCAL-5$^i$. It can be found the confusion problems between novel classes and background are significantly alleviated, which {verifies} the effectiveness of our method.}

\subsection{Ablations}
\textbf{Components analysis.}
BCPT contains two major components, $i.e.$, 
background mining with cluster centers (BMC) and online clustering with guidance (OCG). We validate the effectiveness of each component in Table~\ref{tab:ab_comp}. BMC, which is the key component in BCPT, brings 1.55\% mIoU improvement. OCG contributes extra 1.06\% mIoU. Integrating all components, BCPT significantly improves mIoU from 68.60\%  to 71.21\%.

\noindent \textbf{Number of cluster centers.} We ablate the number of cluster centers and report the results in Table~\ref{tab:ablation2}. The increase of $K$ slightly improves the performance. One reason lies in that the intra-class variance of background can be better modeled with more cluster centers. We set $K=6$ by default.

\section{Conclusion}

{The vast majority of recent efforts in this field seek to achieve improved generalization by fully exploiting the limited supports and proposing advanced meta-training frameworks. Instead, this paper addresses FSS from a new perspective via analyzing the suitability of current pre-training scheme. This leads to the proposed background clustering pre-training for few-shot segmentation (BCPT). To solve the merged background problem, BCPT decouples the novel classes from background via exploiting the underlying semantic structure, which is achieved by an online clustering strategy and learning the embeddings with multiple cluster centers. BCPT further utilizes base classes to guide the clustering process to achieve more accurate clustering results.
BCPT is verified to be effective and sets the new SOTA on PASCAL-5$^i$. In future, we plan to explore more efficient and powerful pre-training methods for FSS.}



\vfill\pagebreak



\bibliographystyle{IEEEbib}

 \bibliography{strings,refs}

\begin{thebibliography}{10}

\bibitem{lu2021simpler}
Zhihe Lu, Sen He, Xiatian Zhu, Li~Zhang, Yi-Zhe Song, and Tao Xiang,
\newblock ``Simpler is better: Few-shot semantic segmentation with classifier weight transformer,''
\newblock in {\em ICCV}, 2021, pp. 8741--8750.

\bibitem{lang2022learning}
Chunbo Lang, Gong Cheng, Binfei Tu, and Junwei Han,
\newblock ``Learning what not to segment: A new perspective on few-shot segmentation,''
\newblock in {\em CVPR}, 2022, pp. 8057--8067.

\bibitem{iqbal2022msanet}
Ehtesham Iqbal, Sirojbek Safarov, and Seongdeok Bang,
\newblock ``Msanet: Multi-similarity and attention guidance for boosting few-shot segmentation,''
\newblock {\em arXiv preprint arXiv:2206.09667}, 2022.

\bibitem{ye2020few}
Han-Jia Ye, Hexiang Hu, De-Chuan Zhan, and Fei Sha,
\newblock ``Few-shot learning via embedding adaptation with set-to-set functions,''
\newblock in {\em CVPR}, 2020, pp. 8808--8817.

\bibitem{zhang2020deepemd}
Chi Zhang, Yujun Cai, Guosheng Lin, and Chunhua Shen,
\newblock ``Deepemd: Few-shot image classification with differentiable earth mover's distance and structured classifiers,''
\newblock in {\em CVPR}, 2020, pp. 12203--12213.

\bibitem{siam2019adaptive}
Mennatullah Siam, Boris Oreshkin, and Martin Jagersand,
\newblock ``Adaptive masked proxies for few-shot segmentation,''
\newblock {\em arXiv preprint arXiv:1902.11123}, 2019.

\bibitem{yang2020prototype}
Boyu Yang, Chang Liu, Bohao Li, Jianbin Jiao, and Qixiang Ye,
\newblock ``Prototype mixture models for few-shot semantic segmentation,''
\newblock in {\em ECCV}. Springer, 2020, pp. 763--778.

\bibitem{liu2020part}
Yongfei Liu, Xiangyi Zhang, Songyang Zhang, and Xuming He,
\newblock ``Part-aware prototype network for few-shot semantic segmentation,''
\newblock in {\em ECCV}. Springer, 2020, pp. 142--158.

\bibitem{nguyen2019feature}
Khoi Nguyen and Sinisa Todorovic,
\newblock ``Feature weighting and boosting for few-shot segmentation,''
\newblock in {\em ICCV}, 2019, pp. 622--631.

\bibitem{yang2021mining}
Lihe Yang, Wei Zhuo, Lei Qi, Yinghuan Shi, and Yang Gao,
\newblock ``Mining latent classes for few-shot segmentation,''
\newblock in {\em ICCV}, 2021, pp. 8721--8730.

\bibitem{liu2022dynamic}
Jie Liu, Yanqi Bao, Guo-Sen Xie, Huan Xiong, Jan-Jakob Sonke, and Efstratios Gavves,
\newblock ``Dynamic prototype convolution network for few-shot semantic segmentation,''
\newblock in {\em CVPR}, 2022, pp. 11553--11562.

\bibitem{liu2020dynamic}
Lizhao Liu, Junyi Cao, Minqian Liu, Yong Guo, Qi~Chen, and Mingkui Tan,
\newblock ``Dynamic extension nets for few-shot semantic segmentation,''
\newblock in {\em Proceedings of the 28th ACM international conference on multimedia}, 2020, pp. 1441--1449.

\bibitem{zhang2021few}
Gengwei Zhang, Guoliang Kang, Yi~Yang, and Yunchao Wei,
\newblock ``Few-shot segmentation via cycle-consistent transformer,''
\newblock {\em NeurIPS}, vol. 34, pp. 21984--21996, 2021.

\bibitem{hong2021cost}
Sunghwan Hong, Seokju Cho, Jisu Nam, and Seungryong Kim,
\newblock ``Cost aggregation is all you need for few-shot segmentation,''
\newblock {\em arXiv preprint arXiv:2112.11685}, 2021.

\bibitem{min2021hypercorrelation}
Juhong Min, Dahyun Kang, and Minsu Cho,
\newblock ``Hypercorrelation squeeze for few-shot segmentation,''
\newblock in {\em ICCV}, 2021, pp. 6941--6952.

\bibitem{shi2022dense}
Xinyu Shi, Dong Wei, Yu~Zhang, Donghuan Lu, Munan Ning, Jiashun Chen, Kai Ma, and Yefeng Zheng,
\newblock ``Dense cross-query-and-support attention weighted mask aggregation for few-shot segmentation,''
\newblock {\em arXiv preprint arXiv:2207.08549}, 2022.

\bibitem{caron2018deep}
Mathilde Caron, Piotr Bojanowski, Armand Joulin, and Matthijs Douze,
\newblock ``Deep clustering for unsupervised learning of visual features,''
\newblock in {\em Proceedings of the ECCV (ECCV)}, 2018, pp. 132--149.

\bibitem{zhan2020online}
Xiaohang Zhan, Jiahao Xie, Ziwei Liu, Yew-Soon Ong, and Chen~Change Loy,
\newblock ``Online deep clustering for unsupervised representation learning,''
\newblock in {\em CVPR}, 2020, pp. 6688--6697.

\bibitem{xie2022contrastive}
Jinheng Xie, Jianfeng Xiang, Junliang Chen, Xianxu Hou, Xiaodong Zhao, and Linlin Shen,
\newblock ``Contrastive learning of class-agnostic activation map for weakly supervised object localization and semantic segmentation,''
\newblock {\em arXiv preprint arXiv:2203.13505}, 2022.

\bibitem{shaban2017one}
Amirreza Shaban, Shray Bansal, Zhen Liu, Irfan Essa, and Byron Boots,
\newblock ``One-shot learning for semantic segmentation,''
\newblock {\em arXiv preprint arXiv:1709.03410}, 2017.

\bibitem{caron2019unsupervised}
Mathilde Caron, Piotr Bojanowski, Julien Mairal, and Armand Joulin,
\newblock ``Unsupervised pre-training of image features on non-curated data,''
\newblock in {\em ICCV}, 2019, pp. 2959--2968.

\bibitem{doersch2015unsupervised}
Carl Doersch, Abhinav Gupta, and Alexei~A Efros,
\newblock ``Unsupervised visual representation learning by context prediction,''
\newblock in {\em Proceedings of the IEEE international conference on computer vision}, 2015, pp. 1422--1430.

\bibitem{tian2020prior}
Zhuotao Tian, Hengshuang Zhao, Michelle Shu, Zhicheng Yang, Ruiyu Li, and Jiaya Jia,
\newblock ``Prior guided feature enrichment network for few-shot segmentation,''
\newblock {\em IEEE transactions on pattern analysis and machine intelligence}, 2020.

\bibitem{fan2022self}
Qi~Fan, Wenjie Pei, Yu-Wing Tai, and Chi-Keung Tang,
\newblock ``Self-support few-shot semantic segmentation,''
\newblock in {\em ECCV}. Springer, 2022, pp. 701--719.

\bibitem{deng2009imagenet}
Jia Deng, Wei Dong, Richard Socher, Li-Jia Li, Kai Li, and Li~Fei-Fei,
\newblock ``Imagenet: A large-scale hierarchical image database,''
\newblock in {\em CVPR}. Ieee, 2009, pp. 248--255.

\bibitem{simonyan2014very}
Karen Simonyan and Andrew Zisserman,
\newblock ``Very deep convolutional networks for large-scale image recognition,''
\newblock {\em arXiv preprint arXiv:1409.1556}, 2014.

\bibitem{he2016deep}
Kaiming He, Xiangyu Zhang, Shaoqing Ren, and Jian Sun,
\newblock ``Deep residual learning for image recognition,''
\newblock in {\em CVPR}, 2016, pp. 770--778.

\bibitem{sun2022singular}
Yanpeng Sun, Qiang Chen, Xiangyu He, Jian Wang, Haocheng Feng, Junyu Han, Errui Ding, Jian Cheng, Zechao Li, and Jingdong Wang,
\newblock ``Singular value fine-tuning: Few-shot segmentation requires few-parameters fine-tuning,''
\newblock {\em arXiv preprint arXiv:2206.06122}, 2022.

\end{thebibliography}

\end{document}